\documentclass[conference]{IEEEtran}
\IEEEoverridecommandlockouts
\usepackage{cite}
\usepackage{amsmath,amssymb,amsfonts}
\usepackage{algorithmic}
\usepackage{graphicx}
\usepackage{textcomp}
\usepackage{xcolor}
\usepackage{float}
\def\BibTeX{{\rm B\kern-.05em{\sc i\kern-.025em b}\kern-.08em
    T\kern-.1667em\lower.7ex\hbox{E}\kern-.125emX}}
\begin{document}

\title{Predicting Brain Degeneration with a Multimodal Siamese Neural Network*\\
{\footnotesize \textsuperscript{*}for  the  Alzheimer’s  Disease  Neuroimaging  Initiative}
\thanks{Data used in preparation of this article were obtained from the Alzheimer’s Disease Neuroimaging Initiative (ADNI)  database  (adni.loni.usc.edu).  As  such,  the  investigators  within  the  ADNI  contributed  to  the  design and implementation of ADNI and/or provided data but did not participate in analysis or writing of this report. A complete listing of ADNI investigators can be found at: http://adni.loni.usc.edu/wp-content/uploads/how\_to\_apply/ADNI\_Acknowledgement\_List.pdf \\

978-1-7281-8750-1/20/\$31.00~\copyright2020  IEEE. Personal use of this material is permitted.  Permission from IEEE must be obtained for all other uses, in any current or future media, including reprinting/republishing this material for advertising or promotional purposes, creating new collective works, for resale or redistribution to servers or lists, or reuse of any copyrighted component of this work in other works. }}

\author{\IEEEauthorblockN{1\textsuperscript{st} Cecilia Ostertag}
\IEEEauthorblockA{\textit{LaRochelle University} \\
\textit{L3i - EA 2118}\\
LaRochelle, France \\
cecilia.ostertag1@univ-lr.fr}
\and
\IEEEauthorblockN{2\textsuperscript{nd} Marie Beurton-Aimar}
\IEEEauthorblockA{\textit{Bordeaux University} \\
\textit{LaBRI -
  CNRS 5800}\\
Bordeaux, France \\
beurton@labri.fr}
\and
\IEEEauthorblockN{3\textsuperscript{rd} Muriel Visani}
\IEEEauthorblockA{\textit{LaRochelle University} \\
\textit{L3i - EA 2118}\\
LaRochelle,
  France \\
muriel.visani@univ-lr.fr}
\and
\IEEEauthorblockN{4\textsuperscript{th} Thierry Urruty}
\IEEEauthorblockA{\textit{Poitiers University} \\
\textit{XLIM CNRS 7252}\\
Poitiers, France \\
thierry.urruty@univ-poitiers.fr}
\and
\IEEEauthorblockN{5\textsuperscript{th} Karell Bertet}
\IEEEauthorblockA{\textit{LaRochelle University} \\
\textit{L3i - EA 2118}\\
LaRochelle,
  France \\
karell.bertet@univ-lr.fr}
}

\IEEEoverridecommandlockouts

\IEEEpubid{\makebox[\columnwidth]{978-1-7281-8750-1/20/\$31.00~\copyright2020 IEEE. } \hspace{\columnsep}\makebox[\columnwidth]{ }}

\maketitle

\IEEEpubidadjcol

\begin{abstract}
To study neurodegenerative diseases, longitudinal studies are carried on volunteer patients. During a time span of several months to several years, they go through regular medical visits to acquire data from different modalities, such as biological samples, cognitive tests, structural and functional imaging. These variables are heterogeneous but they all depend on the patient's health condition, meaning that there are possibly unknown relationships between all modalities. Some information may be specific to some modalities, others may be complementary, and others may be redundant. Some data may also be missing. In this work we present a neural network architecture for multimodal learning, able to use imaging and clinical data from two time points to predict the evolution of a neurodegenerative disease, and robust to missing values. Our multimodal network achieves 92.5\% accuracy and an AUC score of 0.978 over a test set of 57 subjects. We also show the superiority of the multimodal architecture, for up to 37.5\% of missing values in test set subjects' clinical measurements, compared to a model using only the clinical modality.
\end{abstract}

\section{Introduction}
Neurodegenerative diseases are characterized by brain tissue damage and cognitive decline. For example in Alzheimer's disease, patients' brains show ``degenerative changes in selected brain regions, including the temporal and parietal lobes and restricted regions within the frontal cortex and cingulate gyrus" \cite{wenk2003neuropathologic}, and clinical symptoms include short-term memory loss and dysfunctionment of executive functions. The evolution of these symptoms are monitored by clinical measures, e.g. neuropychological tests such as the Mini Mental State Evaluation (MMSE) \cite{folstein1975mini}. Morphological changes can be studied with medical imaging techniques such as Computed Tomography (CT), Magnetic Resonance Imaging (MRI), or Positon Emission Tomography (PET). 

We designed a Deep Neural Network model with sub-modules adapted to different modalities, namely 3D brain MR images and clinical measures. To predict the evolution of brain degeneration, this model uses data from two medical visits. To test our model, we used publicly available data from the Alzheimer's Disease Neuroimaging Initiative (ADNI) database \cite{noauthor_adni_nodate}. Using a subset of subjects from this database, we evaluate the added value of a multimodal approach compared to the use of a model using only clinical attributes. We also compare the robustness of these two models to an increasing proportion of missing values in the test set, and finally we propose a training strategy to learn to handle missing data during training. 

\section{Related Works}

Data acquired from medical visits have different characteristics depending on their modalities. The first step to interpret them is to extract information from each modality, then reduce them to a common space, and finally merge them before the classification step. The question of how and when the fusion of modalities should occur is discussed in Ramachandram et. al.'s review \cite{ramachandram2017deep}, focusing on the distinction between early fusion by concatenation or Principal Component Analysis (PCA) before the model, late fusion by pooling individual decisions from modality-specific models, and intermediate fusion with Deep Neural Networks (DNN).

Several biomedical studies use different imaging modalities, but only a few use both imaging and clinical data. Multi-kernel Support Vector Machines (SVM) \cite{freund1995desicion} ar used by \cite{hinrichs2011predictive} and \cite{eshaghi2015classification} to learn from brain morphometry measures and cognitive scores. Using multi-kernel SVM makes it easy to interpret the influence of each modality on the classification result, because the kernel weights vary according to their relative importance. However, the kernels are combined linearly, so it is not possible to infer correlations between modalities. Random forests (RF) have also been used in \cite{galveia2018ophthalmology} to handle imaging data, free text, and demographic information, after feature extraction with 2D Convolutional Neural Networks (CNN) and a medical ontology. Finally, DNNs are increasingly popular for multimodal learning, as they were used to study Alzheimer's disease \cite{bhagwat2018modeling,lu2018multimodal,lee2019predicting}, amyotrophic lateral sclerosis \cite{van2017deep}, and cervical dysplasia \cite{xu2016multimodal}. DNNs are mathematical models based on elementary blocks called artificial neurons. They are trained iteratively on a dataset, using a loss function to evaluate the error rate of the model at each iteration. At each iteration, the neuron weights are adjusted using the gradient backpropagation algorithm. There are different types of DNNs, depending on the input data or on the task. Vectorial data is usually processed by Feed Forward Neural Networks, also called Multi Layer Perceptrons (MLP), where every neuron in a layer is connected to all neurons in the next layer. For images, CNNs are better suited, thanks to their ability to take into account spatial information \cite{lecun1995comparison}. Multimodal DNN architectures have specific parts to learn independantly from each modality, followed by a part dedicated to modality fusion.

\section{Proposed approach}

    \subsection{Ground Truth Labelling}
        
        Subjects from the ADNI study were followed during three years, with regular 6-monthly medical examinations consisting of PET scans, MRI, biological measures and cognitive examinations. To create labels corresponding to Alzheimer's Disease evolution, we used a method proposed by Bhagwat et. al. \cite{bhagwat2018modeling}: the MMSE score is used to perform a hierarchical clustering on ADNI subjects based on its evolution during the three-years study. This analysis showed two groups of individuals : a group who had stable cognitive scores overtime, and another group whose cognitive scores decreased significantly. We were then able to label the subjects respectively as cognitively stable (class ``Stable") or cognitively declining (class ``Decline"), based on their proximity to each of the two groups, and using MMSE scores from at least 3 timepoints obtained in a timespan longer than one year. We filtered our dataset to keep only ADNI subjects who had the baseline and 12-month MRI. Then, to increase the size of our dataset, we also added subjects with baseline and 6-month MRI. Finally, we got a dataset of 377 patients (191 stable and 186 in cognitive decline).

\subsection{Network Architecture}
    
\begin{figure*}[h]
        \begin{center}
        \includegraphics[width=\textwidth]{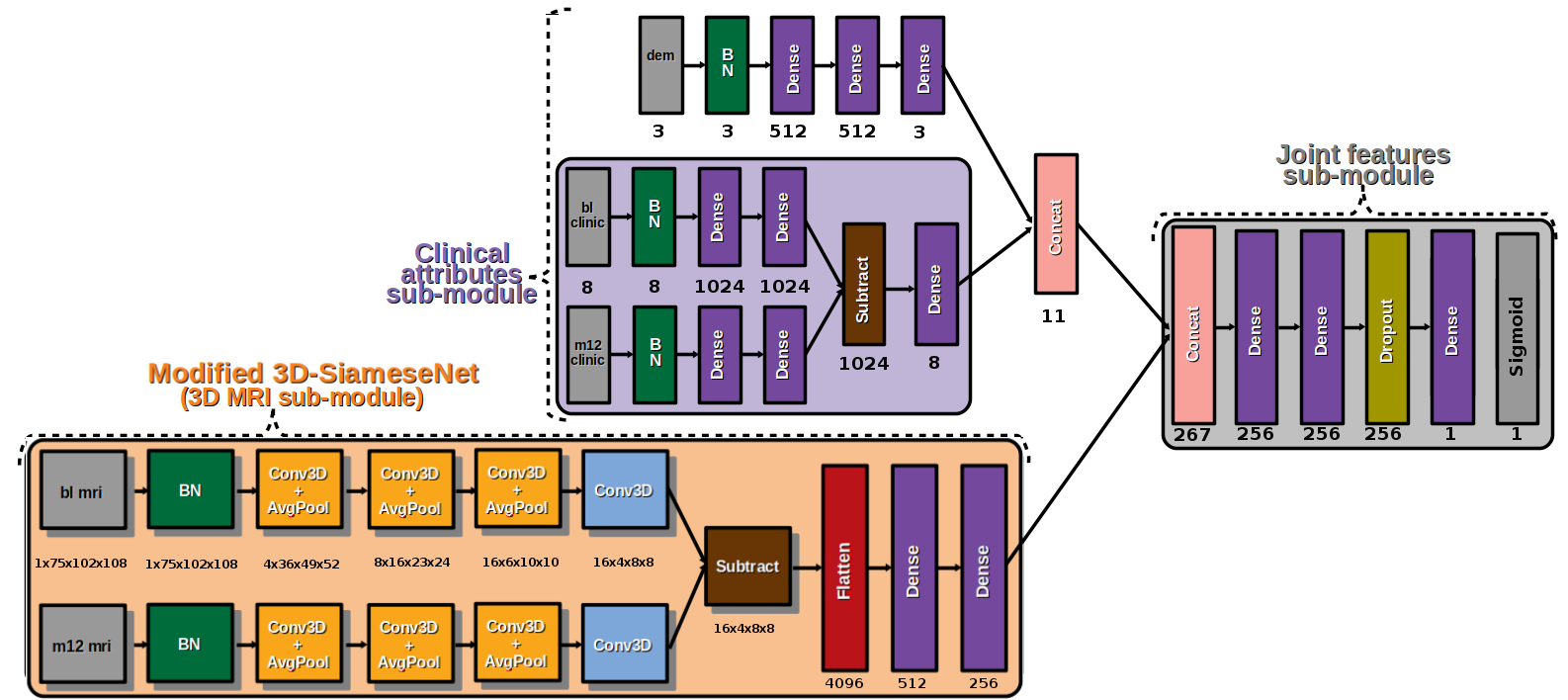}
        \caption{Simplified architecture of our multimodal network. The ``Clinical attributes sub-module" is the base of our ``ClinSiamNN" model, the only change being the addition of a Dropout layer and a Sigmoid activation for prediction. ``bl": Baseline, ``m12": 12-month follow-up (can be replaced by 6-month follow-up)} 
        \label{fig-multi}
        \end{center} 
        \end{figure*}

        Our proposed architecture (see Fig.\ref{fig-multi}) is a complex DNN with modality-specific sub-modules. Each sub-module is itself a neural network, with the type of layers and parameters chosen to extract each modality's features in the most efficient way. Since we are dealing with longitudinal data, our model must take into account its temporality. A common way to do this is to use Recurrent Neural Networks (RNN) to obtain a temporal representation of the data evolution. This approach was used in \cite{lee2019predicting} for the detection of subjects converting to Alzheimer's disease. Their model gave good results, but we note two weaknesses: firstly there are no additional layers after modality fusion, which means that correlations between modalities are not learned by their model, and secondly the authors use morphometric measures computed from MR and PET images instead of the whole brain images. Another method was used by \cite{bhagwat2018modeling}, who employed Siamese Networks to use data (ROI measures from MRI, and cognitive scores) from two medical visits instead of the whole time series. Siamese networks are DNNs made of two identical branches, that share the same weights at initialization and during training. They are usually used to compare pairs of data, for example in the case of face recognition \cite{koch2015siamese,lin2016homemade,zagoruyko2015learning}. 
        
        Given that we want to use the 3D brain images as input to our model, the use of a recurrent model would require too much memory and computation time. This is why we use the siamese network approach to compute a high level representation of changes in a subject's data between two time-points. MRI images, reflecting the subjects' brain morphology which is susceptible to vary during the follow-up study, are fed to a 3D Convolutional Siamese Network, to compute features describing brain degeneration. Clinical scores, also susceptible to vary during the study, are fed to a Feed Forward Siamese Network, to compute features describing the clinical symptoms of cognitive decay. Demographics and genetic data are fed to a simple Feed Forward Network. For training we used pairs of medical visits for each subject.

        \subsubsection{3D Convolutional Siamese Network for Imaging Data}

        \begin{figure}[H]
        \centering
        \includegraphics[scale=0.2]{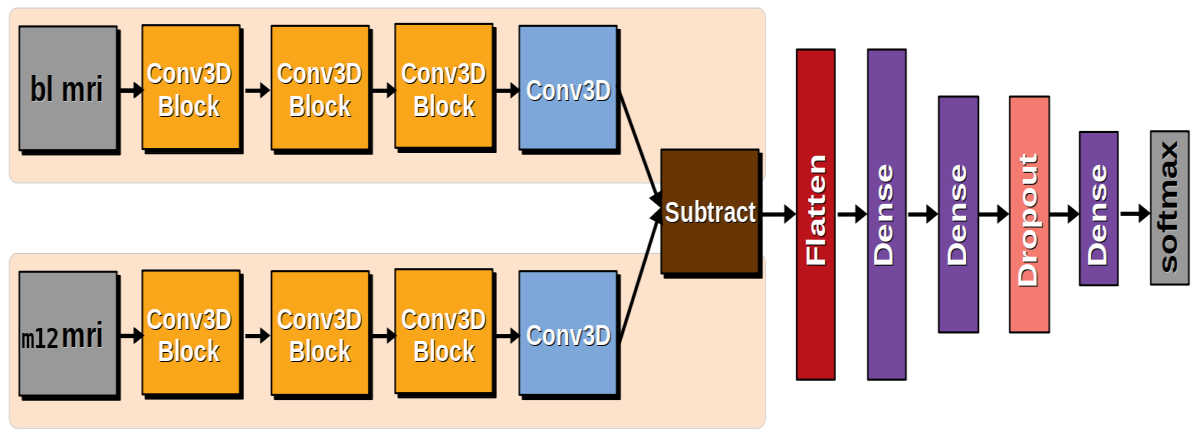}
        \caption{Simplified architecture of our 3D-SiameseNet \cite{prevwork}}
        \label{fig-3dsnet}
        \end{figure}
        
        The MRI sub-module is a slightly modified version of our previous architecture, 3D-SiameseNet \cite{prevwork} (see Fig.\ref{fig-3dsnet}), a Siamese network in which each branch is made of 3D convolution layers followed by average pooling. It is designed to extract features from whole-brain MRI images using all three dimensional context available, and to compute abstract features representing brain evolution between the baseline visit and a follow-up visit from the same subject. It achieved 90\% accuracy for the classification of ``Stable" versus ``Decline" ADNI subjects. The MRI sub-module has less filters in Conv3D layers, to reduce computation and memory costs.

        \subsubsection{Feed Forward Siamese Network for Clinical Data}
        
        As clinical data are a small list of numerical values, we use a Siamese architecture in which each branch is made of fully-connected layers. Similarly to the MRI sub-module, the baseline and follow-up clinical features are merged by a Subtract layer. Since demographics and genetics will either not vary or vary regardless of the disease, they are fed to a simple sequence of fully-connected layers, and later concatenated with the output of the remaining clinical attributes sub-module (see Fig.\ref{fig-multi}). In the following parts of this work, we will compare the performance of this ``Clinical model" (classifier made of the clinical sub-module and the addition of a Dropout layer followed by a Sigmoid activation) to the full ``Multimodal model".
        
        \subsubsection{Joint-features Layers to Merge Modalities}
        
        Once imaging and clinical features are extracted by their respective sub-modules, we use an intermediate fusion strategy by concatenating the two outputs and feeding them to another series of fully-connected layers. This ``joint-features" sub-module is intended for the network to learn correlations between the modalities \cite{xu2016multimodal}. Finally we use a dropout layer with a 0.5 rate, and a final dense layer with a Sigmoid activation to predict the two classes, ``Stable" and ``Decline".

\subsection{Dataset Construction}
        
        \subsubsection{MRI Pre-processing and Augmentation}
        
        Because of memory constraints, our original MRI images were downscaled by 2 to obtain smaller volumes of size 102$\times$108$\times$75 pixels. Moreover, the brain images were skull-stripped \cite{skull} to remove features non related to brain morphology, and their histograms were scaled between 0 and 1. To augment the size of our dataset during training, we computed random left-right rotations between 0 and 5 degrees, Gaussian blur with random standard deviation between 0 and 0.8, and random contrast stretching. These augmentation strategies were chosen amongst popular data augmentation techniques \cite{shorten2019survey}, with the aim of creating plausible brain MR images.

        \subsubsection{Clinical Data Pre-processing}
        
        A large number of clinical variables are available in the ADNI dataset. They consist of demographics, clinical scores, genetic information, and other biological measures. We computed Pearson correlations between variables, to remove some highly correlated variables. Amongst these variables, we removed those with a high number of missing values. This gives us the following list of 11 clinical attributes : AGE, GENDER, LDELTOTAL, RAVLT learning, RAVLT immediate, APOE4 alleles, CDRSB, FAQ, TRABSCOR, RAVLT forgetting, and DIGITSCOR.

\section{Experiments and Results}

    \subsection{Added Value of multimodality}

    We implemented our models using PyTorch \cite{paszke2019pytorch}. They were trained for 75 epochs using a 4-fold stratified cross-validation strategy, while leaving out 57 test subjects (29 ``Stable" and 28 ``Decline"). To evaluate the added value of multi-modality versus single-modality, we trained our two architectures (``ClinSiamNN": Clinical model, and ``MultiSiamNN": Multimodal model) using the same data and same protocols for cross-validation and testing. 

After training, we used our 57 test subjects to compare the two architectures using different metrics: Area Under ROC curve (AUC), precision, recall, F1-score, and accuracy. In our case, the positive class is ``Decline" and the negative class is ``Stable". Our results (see Table \ref{tab-results}) show that the multimodal network is a better classifier than the clinical network, regardless of the metric used for comparison. We also compared our model with the RNN model presented in \cite{lee2019predicting}. For this comparison, we used 1426 subjects from ADNI-1. Given the expected inputs for this model, we used:
\begin{itemize}
\item all of our selected cognitive scores (LDELTOTAL, RAVLT learning, RAVLT immediate, CDRSB, FAQ, TRABSCOR, RAVLT forgetting, and DIGITSCOR)
\item all 7 measures computed from MR images and available in the ADNI database
\item AGE, GENDER and APOE4 alleles
\end{itemize}
We show that our model using siamese networks and two medical visits gives better results than the RNN models using either three or four successive visits. 
 
 \begin{table}[h]
\centering
\caption{Accuracy (Acc), precision (Pre), recall (Rec), Area Under ROC Curve (AUC), and F1-score (F1) obtained with \cite{lee2019predicting} model (1142 train subjects, 284 test subjects), and  our clinical and multimodal models (320 train subjects, 57 test subjects)}
\begin{tabular}{|l|c|c|c|c|c|c|c|c|}
\hline
\multicolumn{1}{|c|}{\textbf{Model}} &  \multicolumn{1}{l|}{\textbf{Time steps}} & \textbf{Acc} & \textbf{Pre} & \textbf{Rec} & \textbf{AUC}   & \textbf{F1}    \\ \hline
RNN \cite{lee2019predicting}                                          & bl to m12                                & 0.846             & 0.807              & 0.840           & 0.918          & 0.822          \\ \hline
RNN \cite{lee2019predicting}                                           & bl to m18                                & 0.853             & 0.819              & 0.839           & 0.918          & 0.829          \\ \hline
ClinSiamNN                                       & bl + m06/12                          & 0.899             & 0.822              & 0.92            & 0.968          & 0.899          \\ \hline
MultiSiamNN                                         & bl + m06/12                          & 0.925             & \textbf{0.924}     & \textbf{0.929}  & \textbf{0.978} & \textbf{0.925} \\ \hline
\end{tabular}
\label{tab-results}
\end{table}

With a dataset of 1116 ADNI subjects, Bhagwat et. al.'s multimodal model had obtained 0.94 accuracy and 0.99 AUC \cite{bhagwat2018modeling}, however we must note that they included the MMSE score used to create the ground truth labels as a clinical variable fed to their model, which may introduce a bias leading to a higher classification performance. 

\subsection{Robustness to Missing Values}

In longitudinal clinical studies, there is always a risk to end up with incomplete data at the end of the study. Concerning this issue, Little and Rubin defined three types of missing data \cite{little2019statistical} :
\begin{itemize}
\item Missing Completely At Random (MCAR), when the probability of having a missing data point is not related to an observed or unobserved variable (Ex: failure of an equipment)
\item Missing At Random (MAR), when the probability of having a missing data point can be related to observed or unobserved variables but not to the trial outcome itself (Ex: subject dropout based on known baseline characteristics)
\item Missing Not At Random (MNAR), when the probability of having a missing data point is related to the trial outcome (Ex: subject not healthy enough to go to the visit)  
\end{itemize}
When facing this problem, a strategy is to filter the dataset and keep only subjects with complete data (Complete Case Analysis), even if this means removing a large part of the dataset. Another common strategy is to replace the missing values using various imputation methods : single or multiple imputation, likelihood-based analysis, or per-protocol analysis \cite{dziura2013strategies}. However, few approaches focus on the architecture of the models themselves. In \cite{smieja2018processing}, the authors introduce an input layer able to replace missing values by a probability density function. Here we decided to test whether the addition of brain MRI data as a second modality could compensate for MCAR data in the clinical measures.

\begin{figure}[!t]
  \centering
  \begin{minipage}[b]{0.45\textwidth}
    \includegraphics[width=\textwidth]{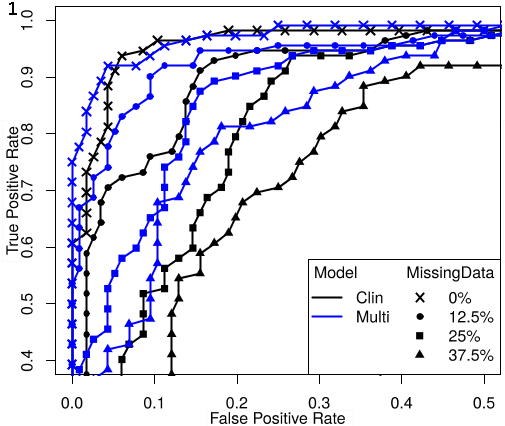}
    \caption{ROC curves of our ``Clinical" and ``Multimodal" models, obtained with 57 test subjects}
    \label{fig-roc1}
  \end{minipage}
  \begin{minipage}[b]{0.45\textwidth}
    \includegraphics[width=\textwidth]{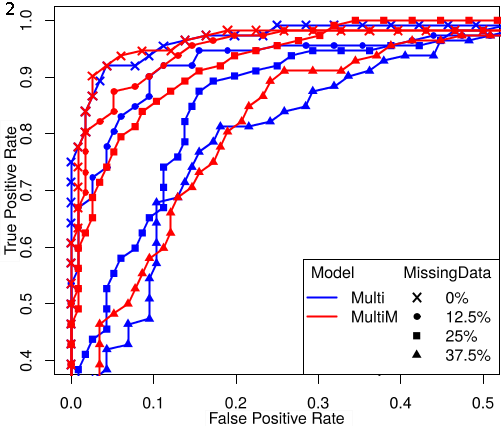}
    \caption{ROC curves of our ``Multimodal" and ``Multimodal trained with NA" models (see section IV.C.)}
    \label{fig-roc2}
  \end{minipage}
\end{figure}

\begin{table*}[h]
\centering
\caption{Area Under ROC Curve (AUC) values for a proportion of missing data ranging from 0 to 37.5\%, obtained with a test set of 57 patients (see section 4.3. for explanations about the MultiM column)}
\begin{tabular}{|c|c|c|c|c|c|c|c|c|}
\hline
\textbf{MissingData} & \textbf{Clin} & \textbf{Lin.SVM} & \textbf{Rbf.SVM} & \textbf{RF} & \textbf{MLP} & \textbf{Adaboost} & \textbf{Multi} & \textbf{MultiM} \\ \hline
0\%               & 0.968         & 0.860            & 0.856            & 0.867       & 0.863        & 0.865             & \textbf{0.978} & \textbf{0.972}  \\ \hline
12.5\%            & 0.931         & 0.830            & 0.803            & 0.840       & 0.828        & 0.843             & \textbf{0.951} & \textbf{0.964}  \\ \hline
25\%              & 0.876         & 0.770            & 0.742            & 0.783       & 0.769        & 0.786             & \textbf{0.916} & \textbf{0.958}  \\ \hline
37.5\%            & 0.807         & 0.728            & 0.704            & 0.738       & 0.727        & 0.732             & \textbf{0.873} & \textbf{0.887}  \\ \hline
\end{tabular}
\label{tab-auc}
\end{table*}

We used our test set to plot Receiver Operating Characteristic (ROC) curves of our models, for a proportion of missing values (NA) ranging from 0 to 37.5\% in the baseline and follow-up clinical data for each patient (see Fig.\ref{fig-roc1}). Those missing values were manually created, by deleting random values according to given NA percentages. If we compare the ROC curves for the clinical model and the multimodal model, we can see that the performance of the Clinical model drops quickly with the increasing proportion of missing values, while the performance of the Multimodal model drops more slowly. 

The AUC values (see Table \ref{tab-auc}) show that our two models outperform state-of-the art models (Linear and Rbf SVM, RF, MLP, and Adaboost \cite{vapnik1997support} trained with only clinical data), in particular the multimodal network has a superior AUC with 37.5\% of NA (0.873) than the state-of-the art models with 0\% of NA. Finally, a Kruskal-Wallis test on AUC values showed a significant difference between the clinical model and multimodal model at the p $<$ 0.05 level, for 37.5\% of NA (p-value = 0.043).

    \subsection{Learning to Handle Missing Values Through Training}

The ability of a deep learning model to generalize well to new data is strongly dependant on the diversity in the training dataset. Following this idea, we wanted to know whether simply adding random missing values during training could help our multimodal network to be more robust to missing values at inference time. To do this, we trained the same multimodal architecture from scratch, this time randomly removing values in the training set. We implemented a random destruction of baseline and follow-up clinical attributes. At every epoch, each subject's clinical attributes have a 10\% chance of being altered. If this is the case, two random variables in the baseline data and two random variables in the follow-up data are erased and replaced by 0 to simulate missing values. This second multimodal model is called ``MultiM" in the results of Fig.\ref{fig-roc2} and Table \ref{tab-auc}. We can see that the strategy of training the multimodal architecture with randomly missing data increases its performance. Kruskal-Wallis tests on the results of Table \ref{tab-auc} showed significant differences over the range of missing data percentages, at the p $<$ 0.05 level, for the clinical model (p-value = 0.0037) and for the multimodal model trained on complete data (p-value = 0.0338) but not for the multimodal model trained with missing values(``MultiM": p-value = 0.0637).

    \subsection{Discussion}

According to our results, the classifier based on a multimodal architecture has a higher precision and recall than the model based on clinical data only. It is the case whether the data from the test subjects is complete or not. This shows the added value of using the additional information provided by the imaging modality. 

Our model, using only data from two time points, outperforms a model based on Recurrent Neural Networks. This can be attributed to our use of whole brain images instead of ROI measures, but also shows that modeling the evolution of patients from two medical visits can be sufficient to predict their evolution. 

Finally, the ability to handle missing data can be significantly improved by simulating missing values at the training step (Kruskal-Wallis test, p-value $<$ 5\%). Another approach to address the issue of missing data could be to encode missing data points differently according to their type (MCAR, MAR, and MNAR), in what could be a ``meta-data" associated to each value.

It should also be noted that, while our ground truth for cognitive decline was inferred from the evolution of the MMSE score only, in a real life application this decline would be diagnosed by a specialist using a composite of several scores as well as context from the patients' history, and would be a more accurate ground truth. However, the lack of availability of medical experts for dataset labelling is a consistent problem when it comes to bio-medical data.

\section{Conclusion}

In this work, we presented a binary classifier based on a multimodal neural network, with specialized sub-modules for each modality and for modality fusion, able to identify patients with cognitive decline. It has a high precision and recall even with proportions of missing values up to 37.5\% in the clinical data. We also showed that simulating missing data at the training step can lead to better results at the inference step. In future works, we would like to improve this architecture by adding a third modality, functional MRI, using a combination of Recurrent Neural Networks and Siamese Networks. It would also be interesting to implement a dynamic architecture, in order to select modality-specific sub-modules depending on which modalities are available for a given subject.

\bibliographystyle{IEEEtran}
\bibliography{IEEEabrv,biblio}

\end{document}